# On the Conditional Independence Implication Problem: A Lattice-Theoretic Approach

**Mathias Niepert**
Department of Computer Science
Indiana University
Bloomington, IN, USA
mniepert@cs.indiana.edu

**Dirk Van Gucht**
Department of Computer Science
Indiana University
Bloomington, IN, USA
vgucht@cs.indiana.edu

**Marc Gyssens**
Department WNI
Hasselt University & Transnational
University of Limburg, Belgium
marc.gyssens@uhasselt.be

## Abstract

A lattice-theoretic framework is introduced that permits the study of the conditional independence (CI) implication problem relative to the class of discrete probability measures. Semi-lattices are associated with CI statements and a finite, sound and complete inference system relative to semi-lattice inclusions is presented. This system is shown to be (1) sound and complete for saturated CI statements, (2) complete for general CI statements, and (3) sound and complete for stable CI statements. These results yield a criterion that can be used to falsify instances of the implication problem and several heuristics are derived that approximate this "lattice-exclusion" criterion in polynomial time. Finally, we provide experimental results that relate our work to results obtained from other existing inference algorithms.

## 1 Introduction

Conditional independence is an important concept in many calculi for dealing with knowledge and uncertainty in artificial intelligence. The notion plays a fundamental role for learning and reasoning in probabilistic systems which are successfully employed in areas such as computer vision, computational biology, and robotics. Hence, new theoretical findings and algorithmic improvements have the potential to impact many fields of research. A central issue for reasoning about conditional independence is the *probabilistic conditional independence implication problem*, that is, to decide whether a CI statement is entailed by a set of other CI statements relative to the class of discrete probability measures. While it remains open whether this problem is decidable, it is known that there exists no finite, sound and complete inference system (Studený [8]). However, there exist finite sound inference systems that have attracted special interest. The most prominent is the *semi-graphoid* axiom system which was introduced as a set of sound inference rules relative to the class of discrete probability measures (Pearl [6]). One of the main contributions of this paper is to extend the semi-graphoids to a *finite* inference system, denoted by $\mathcal{A}$, which we will show to be (1) sound and complete for saturated CI statements, (2) complete for general CI statements, and (3) sound and complete for stable CI statements (de Waal and van der Gaag [2]), all relative to the class of discrete probability measures.

The techniques we use to obtain these results are made possible through the introduction of a *lattice-theoretic framework*. In this approach, semi-lattices are associated with conditional independence statements, and $\mathcal{A}$ is shown to be sound and complete relative to certain inclusion relationships on these semi-lattices. To make the connection between this framework and the conditional independence implication problem, we first link the latter to an addition-based version of the problem. In particular, we introduce the *additive implication problem* for CI statements relative to certain classes of real-valued functions and specify properties of these classes that guarantee soundness and completeness, respectively, of $\mathcal{A}$ for the implication problem. Through the concept of *multi-information functions* induced by probability measures (Studený [9]), we link the additive implication problem for this class of functions to the probabilistic CI implication problem.

The combination of the lattice-inclusion techniques and the completeness result for conditional independence statements allows us to derive criteria that can be used to falsify instances of the implication problem. We show experimentally that these criteria, some of which can be tested for in polynomial time, work very effectively, and we relate the experimental results to those obtained from a *racing algorithm* introduced by Bouckaert and Studený [1].

## 2 CI Statements and System $\mathcal{A}$

We define CI statements and introduce the finite inference system $\mathcal{A}$ for reasoning about the conditional independence implication problem. We will often write $AB$ for the union $A \cup B$, $ab$ for the set $\{a, b\}$, and $a$ for the singleton set $\{a\}$ whenever the interpretation is clear from the context. Throughout the paper, $S$ denotes a finite implicit set of statistical variables.

**Definition 2.1.** The expression $I(A, B|C)$ where $A$, $B$, and $C$ are pairwise disjoint subsets of $S$ is called a *conditional independence (CI) statement*. If $ABC = S$ we say that $I(A, B|C)$ is *saturated*. If either $A = \emptyset$ and/or $B = \emptyset$ we say that $I(A, B|C)$ is *trivial*.

| | |
|---|---|
| $I(A, \emptyset\|C)$ | **Triviality** |
| $I(A, B\|C) \to I(B, A\|C)$ | **Symmetry** |
| $I(A, BD\|C) \to I(A, D\|C)$ | **Decomposition** |
| $I(A, B\|CD) \wedge I(A, D\|C)$ | **Contraction** |
| $\quad \to I(A, BD\|C)$ | |
| | |
| $I(A, B\|C) \to I(A, B\|CD)$ | **Strong union** |
| $I(A, B\|C) \wedge I(D, E\|AC) \wedge$ | **Strong** |
| $\quad I(D, E\|BC) \to I(D, E\|C)$ | $\quad$ **contraction** |

Figure 1: The inference rules of system $\mathcal{A}$.

The set of inference rules in Figure 1 will be denoted by $\mathcal{A}$. The *triviality*, *symmetry*, *decomposition*, and *contraction* rules are part of the semi-graphoid axioms (Geiger [6]). *Strong union* and *strong contraction* are two additional inference rules. Note that *strong union* is not a sound inference rule relative to the class of discrete probability measures. The derivability of a CI statement $c$ from a set of CI statements $\mathcal{C}$ under the inference rules of system $\mathcal{A}$ is denoted by $\mathcal{C} \vdash c$. The *closure* of $\mathcal{C}$ under $\mathcal{A}$, denoted $\mathcal{C}^+$, is the set $\{c \mid \mathcal{C} \vdash c\}$.

**Lemma 2.2** (de Waal and van der Gaag [2]). *The inference rule* composition

$I(A, B|C) \wedge I(A, D|C) \to I(A, BD|C)$ **Composition**

*can be derived using* strong union *and* contraction.

## 3 Lattice-Theoretic Framework

First, we introduce the lattice-theoretic framework which is at the core of the theory developed in this paper. The approach we take is made possible through the association of conditional independence statements with semi-lattices. In this section, we prove that inference system $\mathcal{A}$ is sound and complete relative to specific semi-lattice inclusions. This result forms the backbone of our work on the conditional independence implication problem.

### 3.1 Semi-Lattices of CI Statements

Given two subsets $A$ and $B$ of $S$, we will write $[A, B]$ for the lattice $\{U \mid A \subseteq U \ \& \ U \subseteq B\}$. We will now associate semi-lattices with conditional independence statements.

**Definition 3.1.** Let $I(A, B|C)$ be a CI statement. The *semi-lattice* of $I(A, B|C)$ is defined by $\mathcal{L}(A, B|C) = [C, S] - ([A, S] \cup [B, S])$.

We will often write $\mathcal{L}(c)$ to denote the semi-lattice of a conditional independence statement $c$, and $\mathcal{L}(\mathcal{C})$ to denote the union of semi-lattices, $\bigcup_{c' \in \mathcal{C}} \mathcal{L}(c')$, of a set of conditional independence statements $\mathcal{C}$. Using the notion of *witnesses* of a conditional independence statement, we can rewrite the associated semi-lattice as a difference-free union of lattices.

**Definition 3.2.** Let $I(A, B|C)$ be a CI statement. The set of all witness sets of $I(A, B|C)$ is defined as $\mathcal{W}(A, B|C) = \{\{a, b\} \mid a \in A \text{ and } b \in B\}$.

Note that if $I(A, B|C)$ is trivial, then $\mathcal{W}(A, B|C) = \emptyset$.

**Lemma 3.3.** *Let $c = I(A, B|C)$ be a CI statement. Then $\mathcal{L}(c) = \bigcup_{W \in \mathcal{W}(c)} [C, \overline{W}]$.*

**Example 3.4.** Let $S = \{a, b, c, d\}$ and let $I(bc, d|a)$ be a CI statement. Then, $\mathcal{L}(bc, d|a) = [a, S] - ([bc, S] \cup [d, S]) = \{a, ab, ac\}$. Furthermore, $\mathcal{W}(bc, d|a) = \{bd, cd\}$ and, therefore, $\mathcal{L}(bc, d|a) = [a, ac] \cup [a, ab] = \{a, ab, ac\}$, using Lemma 3.3.

### 3.2 Soundness and Completeness of Inference System $\mathcal{A}$ for Semi-Lattice Inclusion

We will prove that system $\mathcal{A}$ is sound and complete relative to semi-lattice inclusion. First, we show that if a CI statement can be derived from a set of CI statements under $\mathcal{A}$, then we have a set inclusion relationship between their associated semi-lattices.

**Proposition 3.5.** *Let $\mathcal{C}$ be a set of CI statements, and let $c$ be a CI statement. If $\mathcal{C} \vdash c$, then $\mathcal{L}(\mathcal{C}) \supseteq \mathcal{L}(c)$.*

*Proof.* We prove the statement for strong contraction. The proofs for the other inference rules in $\mathcal{A}$ are analogous and are omitted. Let $U \in \mathcal{L}(D, E|C)$. Then $U \supseteq C$. If $U \supseteq A$, then $U \in \mathcal{L}(D, E|AC)$. If $U \supseteq B$, then $U \in \mathcal{L}(E, D|BC)$. If $U \not\supseteq A$ and $U \not\supseteq B$, then $U \in \mathcal{L}(A, B|C)$. □

A CI statement can be equivalent to a set of other CI statements with respect to the inference system $\mathcal{A}$. The following definition of a *witness decomposition* of a CI statement is aimed to prove this property.

**Definition 3.6.** The *witness decomposition* of the CI statement $I(A, B|C)$ is defined by $wdec(A, B|C) := \{I(a, b|C) \mid a \in A \text{ and } b \in B\}$.

A useful property of the witness decomposition of a CI statement is that its closure under $\mathcal{A}$ is the same as the closure of the CI statement itself. In addition, the semi-lattice of a CI statement is equal to the semi-lattice of its witness decomposition.

**Proposition 3.7.** *Let $c$ be a CI statement. (1) $\{c\}^+ = wdec(c)^+$; and (2) $\mathcal{L}(c) = \bigcup_{c' \in wdec(c)} \mathcal{L}(c')$.*

*Proof.* To prove the first statement, let $c = I(A, B|C)$ and $I(a, b|C) \in wdec(c)$. Then $I(a, b|C)$ can be derived from $I(A, B|C)$ by applications of the *decomposition* rule. Hence, $wdec(c)^+ \subseteq \{c\}^+$. By Definition 3.6 we know that for every $a \in A$ and for all $b \in B$ one has $I(a, b|C) \in wdec(c)$. By repeatedly applying *composition*, we can infer the CI statement $I(a, B|C)$. Hence, for all $a \in A$, one has $I(a, B|C) \in wdec(c)^+$ and by *symmetry* $I(B, a|C) \in wdec(c)^+$. Again, by applying *composition* repeatedly, we can infer $I(B, A|C)$ and by *symmetry* $I(A, B|C)$. Hence, $\{c\}^+ \subseteq wdec(c)^+$.

To prove the second statement, let $I(a, b|C) \in wdec(c)$ and $W = \{a, b\}$. Then $\mathcal{L}(a, b|C) = [C, \overline{W}]$. The statement now follows directly from Definition 3.6 and Lemma 3.3. □

We are now in the position to prove the main result concerning the soundness and completeness of the inference system $\mathcal{A}$ for semi-lattice inclusion.

**Theorem 3.8.** *Let $\mathcal{C}$ be a set of CI statements, and let $c$ be a CI statement. Then $\mathcal{C} \vdash c$ if and only if $\mathcal{L}(\mathcal{C}) \supseteq \mathcal{L}(c)$.*

*Proof.* We already know by Proposition 3.5 that if $\mathcal{C} \vdash c$ then $\mathcal{L}(\mathcal{C}) \supseteq \mathcal{L}(c)$. We now proceed to show the other direction. Let us denote $wdec(\mathcal{C}) = \bigcup_{c' \in \mathcal{C}} wdec(c')$ and let $I(a, b|C) \in wdec(c)$ with $W = \{a, b\}$. From the assumption $\mathcal{L}(\mathcal{C}) \supseteq \mathcal{L}(c)$ and Proposition 3.7(2) it follows that $\mathcal{L}(\mathcal{C}) \supseteq \mathcal{L}(a, b|C)$ (1). By Proposition 3.7(1) it suffices to show that $I(a, b|C) \in wdec(\mathcal{C})^+$. However, we will prove the stronger statement $\forall V \in [C, \overline{W}] : I(a, b|V) \in wdec(\mathcal{C})^+$ by downward induction on the lattice $[C, \overline{W}]$.

For the base case we need to show that $I(a, b|\overline{W}) \in wdec(\mathcal{C})^+$. By (1) $\overline{W}$ is in $\mathcal{L}(\mathcal{C})$. Hence, by Proposition 3.7(1), there exists a CI statement $I(a, b|C') \in wdec(\mathcal{C})$ such that $\overline{W} \in \mathcal{L}(a, b|C')$. Now, since $C' \subseteq \overline{W}$, we can derive $I(a, b|\overline{W})$ through *strong union*.

For the induction step, let $C \subseteq V \subset \overline{W}$. The induction hypothesis states that for all $V'$ with $V \subset V' \subseteq \overline{W}$ one has $I(a, b|V') \in wdec(\mathcal{C})^+$. By (1) $V$ is in $\mathcal{L}(\mathcal{C})$. Hence, by Proposition 3.7(1), there exists a CI statement $I(a', b'|C') \in wdec(\mathcal{C})$ such that $V \in \mathcal{L}(a', b'|C')$. Since $C' \subseteq V$ we can use *strong union* to derive $I(a', b'|V)$. Let $W' = \{a', b'\}$. Note that $W' \cap V = \emptyset$. We distinguish three cases:

- $W' = W$. Then we are done.

- Exactly one of the two elements in $W'$ is not in $W$. Without loss of generality let this element be $b'$. Then we can use *contraction* on the statements $I(a, b'|V)$ and $I(a, b|Vb')$ (the latter is in $wdec(\mathcal{C})^+$ by the induction hypothesis) to derive $I(a, b'b|V)$, and finally *decomposition* to derive $I(a, b|V)$.

- Both elements in $W'$ are not in $W$. We can use *strong contraction* on the statements $I(a', b'|V)$, $I(a, b|Va')$, and $I(a, b|Vb')$ (the latter two are in $wdec(\mathcal{C})^+$ by the induction hypothesis) to derive $I(a, b|V)$.

This concludes the proof. □

**Example 3.9.** Let $S = \{a, b, c, d\}$, let $\mathcal{C} = \{I(a, b|\emptyset), I(c, d|a), I(c, d|b)\}$ and let $c = I(c, d|\emptyset)$. We can derive $c$ from $\mathcal{C}$ using the inference rule *strong contraction*. In addition, $\mathcal{L}(\mathcal{C}) = \{\emptyset, c, d, cd\} \cup \{a, ab\} \cup \{b, ab\} = \{\emptyset, a, b, c, d, ab, cd\}$ and $\mathcal{L}(c) = \{\emptyset, a, b, ab\}$, and, therefore, $\mathcal{L}(\mathcal{C}) \supseteq \mathcal{L}(c)$.

## 4 The Additive Implication Problem for CI Statements

An important result in the study of the implication problem relative to the class of discrete probability measures was gained by Studený who linked it to an additive implication problem (Studený [9]). More specifically, it was shown that for every CI statement $I(A, B|C)$, a discrete probability measure $P$ satisfies $I(A, B|C)$ if and only if the *multi-information function*[1] $M_P$ induced by $P$ satisfies the equality $M_P(C) + M_P(ABC) = M_P(AC) + M_P(BC)$. Thus, the *multiplication-based* probabilistic CI implication problem was related to an *addition-based* implication problem. It is this duality that is at the basis of the results developed in this section. However, rather than immediately focusing on specific classes of *multi-information functions*, which is what we pursue in Section 6, we first consider the additive implication problem for CI statements relative to arbitrary classes of real-valued functions.

By a *real-valued function*, we will always mean a function $F : 2^S \to \mathbf{R}$, i.e., a function that maps each subset of $S$ into a real number.

**Definition 4.1.** Let $I(A, B|C)$ be a CI statement, and let $F$ be a real-valued function. We say that $F$ *a-satisfies* $I(A, B|C)$, and write $\models_F^a I(A, B|C)$, if $F(C) + F(ABC) = F(AC) + F(BC)$.

---

[1] The multi-information function of a probability measure will be formally defined in Section 6.

Relative to the notion of *a-satisfaction*, we can now define the *additive implication problem* for conditional independence statements.

**Definition 4.2** (Additive implication problem). Let $\mathcal{C}$ be a set of CI statements, let $c$ be a CI statement, and let $\mathcal{F}$ be a class of real-valued functions. We say that $\mathcal{C}$ *a-implies* $c$ relative to $\mathcal{F}$, and write $\mathcal{C} \models_{\mathcal{F}}^{a} c$, if each function $F \in \mathcal{F}$ that *a-satisfies* the CI statements in $\mathcal{C}$ also *a-satisfies* the CI statement $c$.

We now define the notion of density of a real-valued function. The density is again a real-valued function and plays a crucial role in reasoning about additive implication problems.

**Definition 4.3.** Let $F$ be a real-valued function. The *density*[2] of $F$ is the real-valued function $\Delta F$ defined by $\Delta F(X) = \sum_{X \subseteq U \subseteq S} (-1)^{|U|-|X|} F(U)$, for each $X \subseteq S$.

The following relationship between a real-valued function and its *density* justifies the name.

**Proposition 4.4.** *Let $F$ be a real-valued function. Then, for each $X \subseteq S$, $F(X) = \sum_{X \subseteq U \subseteq S} \Delta F(U)$.*

The *a-satisfaction* of a real-valued function for a CI statement can be characterized in terms of an equation involving its density function. This characterization is central in developing our results and is a special case of a more general result by Sayrafi and Van Gucht who used it in their study of the *frequent itemset mining problem* (Sayrafi and Van Gucht [7]).

**Proposition 4.5.** *Let $I(A, B|C)$ be a CI statement and and let $F$ be a real-valued function. Then, $\models_{F}^{a} I(A, B|C)$ if and only if $\sum_{U \in \mathcal{L}(A,B|C)} \Delta F(U) = 0$.*

## 5 Properties of Classes of Functions - Soundness and Completeness

In this section we study properties of classes of real-valued functions that guarantee soundness and completeness of $\mathcal{A}$, respectively, for the additive implication problem. How these results relate to probabilistic conditional independence implication will become clear in Section 7 and Section 8.

### 5.1 Soundness

First, we define the notion of soundness of system $\mathcal{A}$ for a given class of real-valued functions.

**Definition 5.1** (Soundness). Let $\mathcal{F}$ be a class of real-valued functions. We say that $\mathcal{A}$ is *sound* relative to $\mathcal{F}$ if, for each set $\mathcal{C}$ of CI statements and each CI statement $c$, we have that $\mathcal{C} \vdash c$ implies $\mathcal{C} \models_{\mathcal{F}}^{a} c$.

---
[2]What we call the *density* is sometimes referred to as the *Möbius inversion* of a real-valued function.

In order to characterize soundness we introduce the following property of classes of real-valued functions.

**Definition 5.2** (Zero-density property). Let $\mathcal{F}$ be a class of real-valued functions. We say that $\mathcal{F}$ has the *zero-density property* if, for each $F \in \mathcal{F}$, for each CI statement $c$, and for each $U \in \mathcal{L}(c)$, one has that if $\models_{F}^{a} c$, then $\Delta F(U) = 0$.

We can now provide various characterizations of the soundness of inference system $\mathcal{A}$ for the additive implication problem for CI statements.

**Theorem 5.3.** *Let $\mathcal{F}$ be a class of real-valued functions. Then, the following statements are equivalent:*

*(1) Strong union and decomposition are sound inference rules relative to $\mathcal{F}$ for the additive implication problem;*

*(2) $\mathcal{F}$ has the zero-density property; and*

*(3) $\mathcal{A}$ is sound relative to $\mathcal{F}$ for the additive implication problem.*

*Proof.* We first prove that statement (1) implies statement (2). Let $F \in \mathcal{F}$, let $I(A, B|C)$ be a CI statement, and assume $\models_{F}^{a} I(A, B|C)$. We now show that $\Delta F(V) = 0$ for each $V \in \mathcal{L}(A, B|C)$. The proof goes by downward induction on the semi-lattice $\mathcal{L}(A, B|C)$. First, we observe that by Lemma 3.3, $\mathcal{L}(A, B|C) = \bigcup_{W \in \mathcal{W}(A,B|C)}[C, \overline{W}]$. Hence, for the base case we must prove that $\Delta F(\overline{W}) = 0$ for each $W \in \mathcal{W}(A, B|C)$. Let $W = \{a, b\} \in \mathcal{W}(A, B|C)$. $I(a, b|C)$ is derivable from $I(A, B|C)$ using the inference rule *decomposition* and therefore $\models_{F}^{a} I(a, b|C)$. Since *strong union* is assumed sound and $C \subseteq \overline{W}$ it follows that $\models_{F}^{a} I(a, b|\overline{W})$. Since $\mathcal{L}(a, b|\overline{W}) = \{\overline{W}\}$ we can invoke Proposition 4.5 to conclude that $\Delta F(\overline{W}) = 0$. For the induction step, let $V \in \mathcal{L}(A, B|C)$. The induction hypothesis states that $\Delta F(U) = 0$ for all $U \in \mathcal{L}(A, B|C)$ that are strict supersets of $V$. Similar to the base case, we can infer that $\models_{F}^{a} I(A', B'|V)$ with $A'$, $B'$, and $V$ pairwise disjoint, $A' \subseteq A$, $B' \subseteq B$, and $C \subseteq V$. Hence, by Proposition 4.5, $\sum_{U \in \mathcal{L}(A',B'|V)} \Delta F(U) = \Delta F(V) = 0$. Since for all $U \in \mathcal{L}(A', B'|V)$ with $U \neq V$, we have by Proposition 3.5 that $V \subset U \in \mathcal{L}(A, B|C)$ and, thus, $\Delta F(U) = 0$ by the induction hypothesis.

We now prove that statement (2) implies statement (3). Let $\mathcal{C}$ be a set of CI statements, let $c$ be a CI statement, and assume that $\mathcal{C} \vdash c$. Since $\mathcal{F}$ has the zero-density property, we have that for each $F \in \mathcal{F}$, if $\models_{F}^{a} \mathcal{C}$ then for each $U \in \mathcal{L}(\mathcal{C})$, $\Delta F(U) = 0$. From $\mathcal{C} \vdash c$ and Proposition 3.5, we have $\mathcal{L}(\mathcal{C}) \supseteq \mathcal{L}(c)$. Hence, for all $F \in \mathcal{F}$ we have that if $F$ *a-satisfies* every CI statements in $\mathcal{C}$, then $F$ *a-satisfies* $c$. Thus, $\mathcal{C} \models_{\mathcal{F}}^{a} c$.

Finally, statement (1) follows trivially from (3). □

## 5.2 Completeness

As with soundness in Subsection 5.1, we begin with the definition of the notion of *completeness* of inference system $\mathcal{A}$ for a given class of real-valued functions.

**Definition 5.4** (Completeness). Let $\mathcal{F}$ be a class of real-valued functions. We say that $\mathcal{A}$ is *complete* for the additive implication problem for CI statements relative to $\mathcal{F}$ if, for each set $\mathcal{C}$ of CI statements and each CI statement $c$, one has that $\mathcal{C} \models_\mathcal{F}^a c$ implies $\mathcal{C} \vdash c$.

We now introduce certain special real-valued functions that are at the basis of defining a property guaranteeing completeness of system $\mathcal{A}$.

**Definition 5.5.** Let $V \subseteq S$. The *Kronecker-density function* of $V$, denoted $\delta_V$, is the real-valued function such that $\delta_V(V) = 1$ and $\delta_V(X) = 0$ if $X \neq V$. The *Kronecker-induced function* of $V$, denoted $F_V$, is the real-valued function whose density function is the Kronecker density function of $V$, i.e., for each $X \subseteq S$, $F_V(X) = \sum_{X \subseteq U \subseteq S} \delta_V(U)$, for each $X \subseteq S$.

We can now define a property on classes of real-valued functions that we will show to guarantee the completeness of system $\mathcal{A}$ for the additive implication problem.

**Definition 5.6** (Kronecker property). Let $\mathcal{F}$ be a class of real-valued functions, and let $\Omega \subseteq 2^S$. We say that $\mathcal{F}$ has the *Kronecker property* on $\Omega$ if, for each $U \in \Omega$, there exists a $c_U \in \mathbf{R}$ ($c_U \neq 0$), and a set $D_U = \{d_V \in \mathbf{R} \mid V \notin \Omega\}$ such that the following real-valued function is in $\mathcal{F}$:

$$F_{\Omega, c_U, D_U} := c_U F_U + \sum_{\substack{V \subseteq S \\ V \notin \Omega}} d_V F_V.$$

Note that for all $X \in \Omega$, $\Delta F_{\Omega, c_U, D_U}(X) = c_U$ if $X = U$ and $\Delta F_{\Omega, c_U, D_U}(X) = 0$ if $X \neq U$.

Let $\Omega^{(2)}$ be the set of all subsets of $S$ that lack at least two of their elements, i.e., $\Omega^{(2)} = \{V \subset S \mid |V| \leq |S| - 2\}$. We can now prove that the Kronecker property on $\Omega^{(2)}$ implies the completeness of system $\mathcal{A}$.

**Theorem 5.7.** *Let $\mathcal{F}$ be a class of real-valued functions. If $\mathcal{F}$ has the Kronecker property on $\Omega^{(2)}$, then system $\mathcal{A}$ is complete for the additive implication problem for CI statements relative to $\mathcal{F}$.*

*Proof.* Assume that $\mathcal{F}$ has the Kronecker property on $\Omega^{(2)}$ but that $\mathcal{A}$ is not complete. Then there exists a set $\mathcal{C}$ of CI statements and a CI statement $c$ such that $\mathcal{C} \models_\mathcal{F}^a c$ but $\mathcal{C} \not\vdash c$, or, equivalently by Theorem 3.8, $\mathcal{L}(c) \not\subseteq \mathcal{L}(\mathcal{C})$. Let $U \in \mathcal{L}(c) - \mathcal{L}(\mathcal{C})$. $U$ must be an element in $\Omega^{(2)}$ by Lemma 3.3. Since $\mathcal{F}$ has the Kronecker property on $\Omega^{(2)}$, we know that there exists a $c_U \in \mathbf{R}$ ($c_U \neq 0$), and a set $D_U = \{d_V \in \mathbf{R} \mid V \notin \Omega^{(2)}\}$ such that $F_{\Omega^{(2)}, c_U, D_U} \in \mathcal{F}$. By Definition 5.6, $\Delta F_{\Omega^{(2)}, c_U, D_U}(X) = 0$ for all other $X \in \Omega^{(2)}$. From Proposition 4.5 it follows that $\models_{F_{\Omega^{(2)}, c_U, D_U}}^a \mathcal{C}$, but $\not\models_{F_{\Omega^{(2)}, c_U, D_U}}^a c$, a contradiction to $\mathcal{C} \models_\mathcal{F}^a c$. $\square$

The following example demonstrates the zero-density and Kronecker properties.

**Example 5.8.** Let $S = \{a, b, c\}$, let $\mathcal{F}_1 = \{F_\emptyset, F_a, F_b, F_c\}$ and $\mathcal{F}_2 = \{F_x\}$, where the densities for each real-valued function are given by the table in Figure 2. The densities of the remaining subsets of $S$ are assumed to be 0 for each function. Now, $\Omega^{(2)} = \{\emptyset, a, b, c\}$ and, therefore, $\mathcal{F}_1$ has the Kronecker property on $\Omega^{(2)}$ since $F_{\Omega^{(2)}, c_U, D_U} = F_U$ for all $U \in \Omega^{(2)}$, and the zero-density property. $\mathcal{F}_2$ does not have the Kronecker property. It also does not have the zero-density property as $\models_{F_x}^a I(b, c|\emptyset)$ but $\Delta F_x(\emptyset) \neq 0$.

|  | $\emptyset$ | $\{a\}$ | $\{b\}$ | $\{c\}$ |
|---|---|---|---|---|
| $\Delta F_\emptyset$ | 0.1 | 0 | 0 | 0 |
| $\Delta F_a$ | 0 | -0.3 | 0 | 0 |
| $\Delta F_b$ | 0 | 0 | -0.6 | 0 |
| $\Delta F_c$ | 0 | 0 | 0 | 0.9 |
| $\Delta F_x$ | -0.2 | 0.2 | 0.6 | 0.3 |

Figure 2: Densities of several real-valued functions.

## 6 The Conditional Independence Implication Problem

While the theory presented so far has been concerned with the additive implication problem for CI statements, it is also applicable to the conditional independence implication problem. The link between these two problems is made with the concept of *multi-information functions* (Studený [9]) induced by probability measures. In this paper we will restrict ourselves to the class of *discrete* probability measures.

**Definition 6.1.** A *probability model* over $S = \{s_1, \ldots, s_n\}$ is a pair $(dom, P)$, where $dom$ is a domain mapping that maps each $s_i$ to a finite domain $dom(s_i)$, and $P$ is a probability measure having $dom(s_1) \times \cdots \times dom(s_n)$ as its sample space. For $A = \{a_1, \ldots, a_k\} \subseteq S$, we will say that $\mathbf{a}$ is a domain vector of $A$ if $\mathbf{a} \in dom(a_1) \times \cdots \times dom(a_k)$.

In what follows, we will only refer to probability measures, keeping their probability models implicit.

**Definition 6.2.** Let $I(A, B|C)$ be a CI statement, and let $P$ be a probability measure. We say that $P$ *m-satisfies* $I(A, B|C)$, and write $\models_P^m I(A, B|C)$, if for every domain vector $\mathbf{a}$, $\mathbf{b}$, and $\mathbf{c}$ of $A$, $B$, and $C$, respectively, $P(\mathbf{c})P(\mathbf{a}, \mathbf{b}, \mathbf{c}) = P(\mathbf{a}, \mathbf{c})P(\mathbf{b}, \mathbf{c})$.

Relative to the notion of *m-satisfaction* we can now define the *probabilistic conditional independence implication problem*.

**Definition 6.3** (Probabilistic conditional independence implication problem). Let $\mathcal{C}$ be a set of CI statements, let $c$ be a CI statement, and let $\mathcal{P}$ be the class of discrete probability measures. We say that $\mathcal{C}$ *m-implies* $c$ relative to $\mathcal{P}$, and write $\mathcal{C} \models_\mathcal{P}^m c$, if each function $P \in \mathcal{P}$ that *m-satisfies* the CI statements in $\mathcal{C}$ also *m-satisfies* the CI statement $c$. The set $\{c \mid \mathcal{C} \models_\mathcal{P}^m c\}$ will be denoted by $\mathcal{C}^*$.

Next, we define the *multi-information function* induced by a probability measure (Studený [9]), which is based on the Kullback-Leibler divergence (Kullback and Leibler [4]).

**Definition 6.4.** Let $P$ and $Q$ be two probability measures over a discrete sample space. Then, the relative entropy (Kullback-Leibler divergence) $H$ is defined as

$$H(P|Q) := \sum_{\mathbf{x}} \{P(\mathbf{x}) \log \frac{P(\mathbf{x})}{Q(\mathbf{x})},\ P(\mathbf{x}) > 0\},$$

with $\mathbf{x}$ ranging over all elements of the discrete sample space.

**Definition 6.5.** Let $P$ be a probability measure, and let $H$ be the relative entropy. The *multi-information function* $M_P : 2^S \to [0, \infty]$ induced by $P$ is defined as

$$M_P(A) := H(P^A | \prod_{a \in A} P^{\{a\}}),$$

for each non-empty subset $A$ of $S$ and $M_P(\emptyset) = 0$.[3]

The class of multi-information functions induced by the class of discrete probability measures $\mathcal{P}$ will be denoted by $\mathcal{M}$. We can now state the fundamental result of Studený that couples the probabilistic CI implication problem with the additive implication problem for CI statements relative to $\mathcal{M}$.

**Theorem 6.6** (Studený [9]). *Let $\mathcal{C}$ be a set of CI statements and let $c$ be a CI statement. Then, $\mathcal{C} \models_\mathcal{M}^a c$ if and only if $\mathcal{C} \models_\mathcal{P}^m c$.*

## 7 Saturated CI Statements - Soundness and Completeness of $\mathcal{A}$

In this section we show that system $\mathcal{A}$ is sound and complete for the probabilistic CI implication problem for *saturated* CI statements. We recall that a CI statement $I(A, B|C)$ is saturated if $ABC = S$. We begin by showing the following technical lemma.

---
[3]Here, $P^A$ and $P^{\{a\}}$ denote the marginal probability measures of $P$ over $A$ and $\{a\}$, respectively.

**Lemma 7.1.** *The class of multi-information functions $\mathcal{M}$ induced by the class of discrete probability measures has the zero-density property with respect to saturated CI statements.*

*Proof.* We have to show that for each saturated CI statements $c$, for each $M \in \mathcal{M}$, and for each $U \in \mathcal{L}(c)$, if $\models_M^a c$, then $\Delta M(U) = 0$. The semi-graphoid inference rules are sound relative to the class of probability measures. Hence, in particular, by Theorem 6.6, *weak union* is sound relative to $\mathcal{M}$, i.e., $\{I(AD, B|C)\} \models_\mathcal{M}^a I(A, B|CD)$. Let $M \in \mathcal{M}$, let $\Delta M$ be the corresponding density function, and let $\models_M^a I(A, B|C)$ with $ABC = S$. In addition, let $I(A, B|C)$ be non-trivial since the proposition is obviously true for trivial CI statements. We will prove by downward induction on the semi-lattice $\mathcal{L}(A, B|C)$ that $\Delta M(U) = 0$ for each $U \in \mathcal{L}(A, B|C)$. Note that this proof is similar to the proof of Proposition 5.3. (Here, *weak union* is used instead of *decomposition* and *strong union*).

For the base case, we show for each $W \in \mathcal{W}(A, B|C)$ that $\Delta M(\overline{W}) = 0$. Let $W = \{a, b\}$. By repeatedly applying *weak union* we can derive $\models_M^a I(a, b|\overline{W})$ because $ABC = S$. Now, since $\mathcal{L}(a, b|\overline{W}) = \{\overline{W}\}$ we can conclude that $\Delta M(\overline{W}) = 0$.

For the induction step, let $V \in \mathcal{L}(A, B|C)$. The induction hypothesis states that $\Delta M(U) = 0$ for each $U \in \mathcal{L}(A, B|C)$ with $U$ a strict superset of $V$. From the given CI statement $I(A, B|C)$ we can derive, again by *weak union*, $I(A', B'|V)$ with $VA'B' = S$ since $V - C \subseteq AB$. Since $\mathcal{L}(A', B'|V)$ contains only $V$ and strict supersets $V'$ of $V$, with $V' \in \mathcal{L}(A, B|C)$, we can conclude that $\sum_{U \in \mathcal{L}(A', B'|V)} \Delta F(U) = \Delta F(V) = 0$ by the induction hypothesis. $\square$

We are now in the position to prove that inference system $\mathcal{A}$ is sound and complete for the probabilistic implication problem for *saturated* conditional independence statements.

**Theorem 7.2.** *$\mathcal{A}$ is sound and complete for the probabilistic conditional independence implication problem for saturated CI statements.*

*Proof.* The soundness follows directly from Lemma 7.1, Theorem 5.3, and Theorem 6.6. To show completeness, notice that the semi-graphoid axioms are derivable under inference system $\mathcal{A}$. Furthermore, Geiger and Pearl proved that the semi-graphoid axioms are complete for the probabilistic conditional independence implication problem for *saturated* CI statements (Geiger and Pearl [3]). $\square$

## 8 CI Statements - Completeness of $\mathcal{A}$

In this section we will show that inference system $\mathcal{A}$ is complete for the probabilistic conditional independence implication problem. We first prove that $\mathcal{M}$ has the Kronecker property on $\Omega^{(2)}$. To show this, it would be sufficient to construct a set of discrete probability measures whose induced multi-information functions are Kronecker-induced functions. However, instead of taking this route, we pursue a different approach by first focusing on results with respect to *saturated* CI statements. We first need the following simple lemma.

**Lemma 8.1.** *For $U \subseteq S$, $\{X \in \Omega^{(2)} \mid X \supseteq U\} = \bigcup_{\substack{U_1 \cup U_2 = \overline{U} \\ U_1 \cap U_2 = \emptyset}} \mathcal{L}(U_1, U_2 | U)$.*

**Proposition 8.2.** *Let $\mathcal{F}$ be a class of real-valued functions. If $\mathcal{A}$ is sound and complete for the additive implication problem relative to $\mathcal{F}$ for saturated CI statements, then $\mathcal{F}$ has the Kronecker property on $\Omega^{(2)}$.*

*Proof.* If $|S| \leq 1$, then $\Omega^{(2)} = \emptyset$ and the statement follows trivially. Hence, assume that $|S| \geq 2$. Suppose that $\mathcal{A}$ is sound and complete for saturated CI statements but that $\mathcal{F}$ does not have the Kronecker property on $\Omega^{(2)}$. Then there exists a set $U \in \Omega^{(2)}$ such that for each $c_U \in \mathbf{R}$ ($c_U \neq 0$), and for each set $D_U = \{d_V \in \mathbf{R} \mid V \notin \Omega^{(2)}\}$ we have $F_{\Omega^{(2)}, c_U, D_U} \notin \mathcal{F}$. Now, let $\mathcal{C}$ be the set of saturated CI statements

$$\{ I(U, \overline{U}|\emptyset) \} \cup \bigcup_{\substack{U_1 \cup U_2 = U \\ U_1 \cap U_2 = \emptyset}} \{ I(U_1, U_2|\overline{U}) \} \cup$$
$$\bigcup_{v \in \overline{U}} \bigcup_{\substack{V_1 \cup V_2 = \overline{U} - \{v\} \\ V_1 \cap V_2 = \emptyset}} \{ I(V_1, V_2|U \cup \{v\}) \},$$

and let $c$ be the saturated CI statement $I(U_1, U_2|U)$ for some non-empty sets $U_1$ and $U_2$. Notice that such sets exist because $|\overline{U}| \geq 2$. By Lemma 8.1 it is $\mathcal{L}(\mathcal{C}) = \Omega^{(2)} - \{U\}$ and $U \in \mathcal{L}(c) \subseteq \Omega^{(2)}$ and therefore $\mathcal{L}(c) \nsubseteq \mathcal{L}(\mathcal{C})$. Hence, by Theorem 3.8, $\mathcal{C} \nvdash c$. We now show that $\mathcal{C} \models_\mathcal{F}^a c$ to obtain the contradiction to the completeness of $\mathcal{A}$. If there does not exist an $F \in \mathcal{F}$ which *a-satisfies* $\mathcal{C}$ we are done because then $\mathcal{C} \models_\mathcal{F}^a c$ follows trivially. Thus, let $F$ be in $\mathcal{F}$ and assume that $\models_F^a \mathcal{C}$. Since $\mathcal{A}$ is sound relative to $\mathcal{F}$ for saturated CI statements, we know by Theorem 5.3 that $\mathcal{F}$ has the zero-density property. Thus, $\Delta F(X) = 0$ for each $X \in \Omega^{(2)}$ with $X \neq U$. But then $\Delta F(U) = 0$ since otherwise there would exist a $c_U \in \mathbf{R}$, $c_U = \Delta F(U) \neq 0$, and a set $D_U = \{d_V \in \mathbf{R} \mid V \notin \Omega^{(2)}\}$ such that $F_{\Omega^{(2)}, c_U, D_U} = F \in \mathcal{F}$. Hence, $F$ must be a function whose density is zero on every element of $\Omega^{(2)}$. Thus, $\models_F^a c$ and it follows that $\mathcal{C} \models_\mathcal{F}^a c$. □

The completeness of $\mathcal{A}$ for the CI implication problem can now be proved based on the previous results.

**Theorem 8.3.** *$\mathcal{A}$ is complete for the probabilistic conditional independence implication problem.*

*Proof.* We know from Theorem 7.2 that $\mathcal{A}$ is sound and complete relative to $\mathcal{M}$ for saturated CI statements. Now, by Proposition 8.2, $\mathcal{M}$ has the Kronecker property on $\Omega^{(2)}$. Finally, through Theorem 5.7 and Theorem 6.6, the statement follows. □

**Example 8.4.** (Studený [9]) described the following sound inference rule relative to discrete probability measures which refuted the conjecture (Pearl [6]) that the semi-graphoid axioms are complete for the probabilistic CI implication problem:

$I(A, B|CD) \wedge I(C, D|A) \wedge I(C, D|B) \wedge I(A, B|\emptyset) \rightarrow I(C, D|AB) \wedge I(A, B|C) \wedge I(A, B|D) \wedge I(C, D|\emptyset).$

By applying *strong contraction* to the statements $I(A, B|\emptyset), I(C, D|A),$ and $I(C, D|B)$ we can derive the statement $I(C, D|\emptyset)$. All the other statements can be derived using *strong union*.

**Remark 8.5.** The inference system $\mathcal{A}$ without *strong contraction* is *not* complete. The consequence $I(C, D|\emptyset)$ of the clause from Example 8.4 cannot be derived from the antecedents without *strong contraction*.

## 9 Complete Axiomatization of Stable Independence

When new information is available to a probabilistic system the set of associated relevant CI statements changes dynamically. However, some of the CI statements will continue to hold. These CI statements were termed *stable* by de Waal and van der Gaag [2]. A first investigation of their structural properties was undertaken by Matúš who used the term *ascending* conditional independence (Matúš [5]). Every set of CI statements can be partitioned into its *stable* and *unstable* part. We will show that inference system $\mathcal{A}$ is sound and complete for the probabilistic CI implication problem for *stable* conditional independence statements.

**Definition 9.1.** Let $\mathcal{C}$ be a set of CI statements, and let $\mathcal{C}^{SG+}$ be the semi-graphoid closure of $\mathcal{C}$. Then $I(A, B|C)$ is said to be *stable* in $\mathcal{C}$, if $I(A, B|C') \in \mathcal{C}^{SG+}$ for all sets $C'$ with $C \subseteq C' \subseteq S$.

**Theorem 9.2.** *Let $\mathcal{C}_S$ be a set of stable CI statements. Then, $\mathcal{A}$ is sound and complete for the probabilistic conditional independence implication problem for $\mathcal{C}_S$, or, equivalently, $\mathcal{C}_S^* = \mathcal{C}_S^+$.*

*Proof.* The soundness follows from Theorem 5.3 and from *strong union* and *decomposition* being sound inference rules relative to $\mathcal{M}$ for stable CI statements. The completeness follows from Theorem 8.3. □

**Remark 9.3.** The previous result is also interesting with respect to the problem of finding a minimal, non-redundant representation of stable independence relations. Here, lattice-inclusion could aid the lossless compaction of representations of stable CI statements: $\mathcal{L}(\mathcal{C}_S - \{c\}) = \mathcal{L}(\mathcal{C}_S)$ if and only if $c$ is redundant in $\mathcal{C}_S$.

## 10 Falsification Algorithm

Theorem 3.8 and Theorem 8.3 lend themselves to a *falsification algorithm*, that is, an algorithm which can falsify instances of the probabilistic conditional independence implication problem. We consider the following corollary which directly follows from these two results.

**Corollary 10.1.** *Let $\mathcal{C}$ be a set of CI statements, and let $\mathcal{P}$ be the class of discrete probability measures. If $\mathcal{L}(\mathcal{C}) \not\supseteq \mathcal{L}(c)$, then $\mathcal{C} \not\models^m_{\mathcal{P}} c$.*

If the falsified implications were, on average, only a small fraction of all those that are falsifiable, the result would be disappointing from a practical point of view. Fortunately, we will not only be able to show that a large number of implications can be falsified by the "lattice-exclusion" criterion identified in Corollary 10.1, but also that polynomial time heuristics exist that provide good approximations of said criterion.

**Falsification Criterion.** Input: A set of CI statements $\mathcal{C}$ and a CI statement $c$. Test: if $\mathcal{L}(\mathcal{C}) \not\supseteq \mathcal{L}(c)$, return "false", else return "unknown."

**Heuristic 1.** Input: A set of CI statements $\mathcal{C}$ and a CI statement $I(A, B|C)$. Test: if for each $I(A', B'|C') \in \mathcal{C}$ it is $C \not\supseteq C'$, return "false", else return "unknown."

**Heuristic 2.** Input: A set of CI statements $\mathcal{C}$, and a CI statement $I(A, B|C)$. Test: if there exists one $W \in \mathcal{W}(A, B|C)$ such that for all $I(A', B'|C') \in \mathcal{C}$ it is $W \notin \mathcal{W}(A', B'|C')$, return "false", else return "unknown."

It follows from Lemma 3.3 that if one of the two heuristics returns "false," then $\mathcal{L}(\mathcal{C}) \not\supseteq \mathcal{L}(c)$, and therefore $\mathcal{C} \not\models^m_{\mathcal{P}} c$ by Corollary 10.1.

**Example 10.2.** Let $S$ be a finite set, and $A, B, C,$ and $D$ be pairwise disjoint subsets of $S$. The inference rule *intersection*, $I(A, B|DC) \wedge I(A, D|BC) \rightarrow I(A, BD|C)$, is *not* sound relative to the class of discrete probability measures. Heuristic 1 can reject this instance of the implication problem in polynomial time in the size of $S$.

**Remark 10.3.** The falsification criterion leads in fact to a *family* of polynomial time heuristics. While Heuristic 1 checks if the unique *meet* (greatest lower bound) of the semi-lattice $\mathcal{L}(c)$ is not in $\mathcal{L}(\mathcal{C})$ and

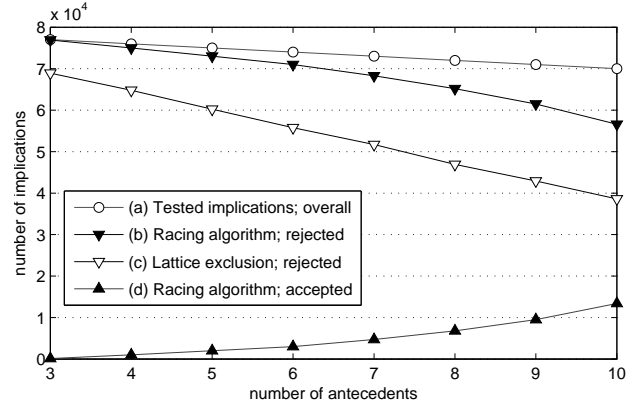

Figure 3: Rejection and acceptance curves of the racing and falsification algorithms, respectively, for five attributes.

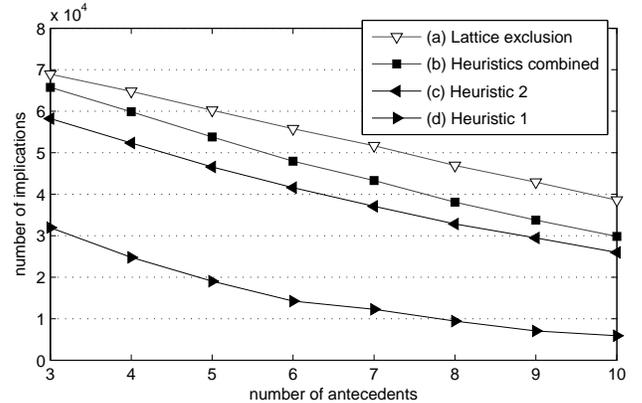

Figure 4: Falsifications based on the lattice-exclusion criterion and the heuristics, for five attributes. The combination of the heuristics reaches 95% of the falsifications of the full-blown lattice exclusion criterion for 3 antecedents down to 77% for 10 antecedents.

Heuristic 2 if the (potentially multiple) *joins* (least upper bounds) of the semi-lattice $\mathcal{L}(c)$ are not in $\mathcal{L}(\mathcal{C})$, we may select additional elements in the semi-lattice $\mathcal{L}(c)$ that are located between these two extrema to derive more falsification heuristics.

With our experiments we want to show that (1) the lattice-exclusion criterion can falsify a large fraction of all falsifiable implications, and (2) that the two provided heuristics are good approximation of the full-blown lattice-exclusion criterion. To make our outcomes comparable to existing results, we adopted the experimental setup for the *racing algorithm* from Bouckaert and Studený [1] (also using 5 attributes). A thousand sets of antecedents each were generated by randomly selecting 3 up to 10 elementary CI statements, resulting in a total of 8000 sets of antecedents.[4]

---
[4]An elementary CI statement is of the form $I(a, b|C)$, where $a, b \in S$ and $C \subseteq S - \{a, b\}$.

The *falsification algorithm* and the heuristics were run on these sets with each of the remaining elementary CI statements as consequence, one at a time. Since there are 80 elementary CI statements for 5 attributes, this resulted in 77000 implication problems for sets with 3 antecedents, 76000 for sets with 4 antecedents, down to 70000 for sets with 10 antecedents.

The rejection procedure of the *racing algorithm* is rooted in the theory of imsets: an instance is rejected if one of the supermodular functions constructed by the algorithm is a counter-model for this instance. It has exponential running time and might reject implications that actually *do* hold. This is a consequence of the fact that $\mathcal{M}$ is a *strict* subset of the class of all supermodular functions. (See Examples 4.1 and 6.2 in Studený's monograph [9].) The *falsification algorithm* based on Corollary 10.1, on the other hand, ensures that if an instance of the implication problem is rejected, then it is guaranteed not to be valid.

Figure 3 shows the rejection curves of the *racing algorithm* (b) and the *falsification algorithm* (c), respectively, and the acceptance curve of the racing algorithm (d). The area between the two rejection curves can be interpreted as the "decision gap", i.e., the amount of instances of the implication problem for which the validity is unknown. The curve marked with circles (a) depicts the total number of tested instances. Figure 4 depicts the rejection curves for the *falsification algorithm* (a), for the combination of Heuristic 1 and Heuristics 2 (b), and for Heuristic 2 (c) and Heuristic 1 (d) run separately. The combination of the heuristics compares favorable with the full-blown *falsification criterion*. The experiments also show that Heuristic 2 is more effective than Heuristic 1.

## 11 Conclusion and Future Work

A complete inference system for the probabilistic conditional independence implication problem was presented and related to the lattice-exclusion criterion. We derived polynomial time approximations that can be used as a preprocessing step to efficiently shrink the search space of possibly valid inferences. We already have experimental evidence that our approach scales to much larger instances of the implication problem than those reported on in this paper. This could, for instance, provide insights into combinatorial bounds for the number of (stable) CI structures. The falsification algorithm and the heuristics can be combined with algorithms that infer valid implications, like the one based on structural imsets which is used as part of the *racing algorithm* [1]. In addition, the lattice exclusion criterion and the heuristics can be utilized to store information about conditional independencies more efficiently, using non-redundant representations. Overall, we believe that the lattice-theoretic framework for reasoning about conditional independence is a novel and powerful tool. We conjecture that there are interesting connections between our theory and Studený's theory of imsets which we will continue to investigate.

**Acknowledgments**

We thank Remco Bouckaert for providing us with the source code of the *racing algorithm*. We also want to thank Milan Studený for the information he provided us concerning the complete axiomatization of the implication problem for four attributes, František Matúš for helpful feedback on an earlier draft, and the anonymous reviewers whose comments helped to improve the quality of the paper.